\documentclass[journal]{IEEEtran}

\usepackage{makeidx}  
\usepackage{graphicx}
\usepackage{amsmath}
\usepackage{gensymb}
\usepackage{amssymb}  
\usepackage[ruled]{algorithm2e}
\usepackage{bm}
\usepackage{caption}
\usepackage{textcomp}
\usepackage{booktabs}
\usepackage{multirow}
\usepackage{url}
\usepackage{nicefrac}
\usepackage{hyperref}
\usepackage{color}
\usepackage[side,hang,flushmargin]{footmisc}
\usepackage[linewidth=1pt,framemethod=TikZ]{mdframed}
\usepackage{float}

\newenvironment{my_mdframed}[1]
  {\mdfsetup{
    frametitle={\colorbox{white}{\space#1\space}},
    innertopmargin=-5pt,
    frametitleaboveskip=-\ht\strutbox,
    frametitlealignment=\center
    }
  \begin{mdframed}%
  }
  {\end{mdframed}}

\newcommand{\textclr}[1]{\textcolor{black}{#1}}

\newcommand\blfootnote[1]{}

\begin{document}

\title{Learning-Based Quality Control for Cardiac MR Images}

\author{Giacomo~Tarroni, Ozan~Oktay, Wenjia~Bai, Andreas~Schuh, Hideaki~Suzuki, Jonathan~Passerat-Palmbach, Antonio~de~Marvao, Declan~P.~O'Regan, Stuart~Cook, Ben~Glocker, Paul~M.~Matthews, Daniel~Rueckert
\thanks{G. Tarroni is with the Department
of Computing, Imperial College London, SW7 2AZ London, UK, e-mail: giacomo.tarroni@gmail.com. O. Oktay, W. Bai, A. Schuh, J. Passerat-Palmbach, B. Glocker and D. Rueckert are also with the Department
of Computing, Imperial College London.}
\thanks{H. Suzuki and P. M. Matthews are with the Division of Brain Sciences, Faculty of Medicine, Imperial College London. P. M. Matthews is also with the UK Dementia Research Institute, London, UK.}%
\thanks{A. de Marvao, D. P. O'Regan and S. Cook are with the MRC London Institute of Medical Sciences, Faculty of Medicine, Imperial College London.}%
\thanks{This work has been submitted to the IEEE for possible publication. Copyright may be transferred without notice, after which this version may no longer be accessible.}
}

\maketitle

\begin{abstract}
The effectiveness of a cardiovascular magnetic resonance (CMR) scan depends on the ability of the operator to correctly tune the acquisition parameters to the subject being scanned and on the potential occurrence of imaging artefacts such as cardiac and respiratory motion. In the clinical practice, a quality control step is performed by visual assessment of the acquired images: however, this procedure is strongly operator-dependent, cumbersome and sometimes incompatible with the time constraints in clinical settings and large-scale studies. We propose a fast, fully-automated, learning-based quality control pipeline for CMR images, specifically for short-axis image stacks. Our pipeline performs three important quality checks: 1) heart coverage estimation, 2) inter-slice motion detection, 3) image contrast estimation in the cardiac region. The pipeline uses a hybrid decision forest method - integrating both regression and structured classification models - to extract landmarks as well as probabilistic segmentation maps from both long- and short-axis images as a basis to perform the quality checks. The technique was tested on up to 3000 cases from the UK Biobank \textclr{\blfootnote{AE.4,\\R2.8}as well as on 100 cases from the UK Digital Heart Project}, and validated against manual annotations and visual inspections performed by expert interpreters. The results show the capability of the proposed pipeline to correctly detect incomplete or corrupted scans \textclr{\blfootnote{R3.1}(e.g. on UK Biobank, sensitivity and specificity respectively 88\% and 99\% for heart coverage estimation, 85\% and 95\% for motion detection)}, allowing their exclusion from the analysed dataset or the triggering of a new acquisition.
\end{abstract}


\IEEEpeerreviewmaketitle

\section{Introduction}

\IEEEPARstart{C}{ardiovascular} magnetic resonance (CMR) imaging presents a wide variety of different applications for the anatomical and functional assessment of the heart. The success of a CMR acquisition relies, however, on the ability of the MR operator to correctly tune the acquisition parameters to the subject being scanned \cite{Zhuo2006}. Moreover, CMR can be negatively affected by a long list of imaging artefacts (caused for instance by respiratory and cardiac motion, blood flow and magnetic field inhomogeneities) \cite{Ferreira2013}. Therefore, a quality control step is required to assess the usability of the acquired images. In the clinical practice this step is performed by visual inspection, usually carried out by the same operator who set up the acquisition, thus leading to highly subjective results.
\begin{figure}[!t]
\centering
\includegraphics[width=\columnwidth]{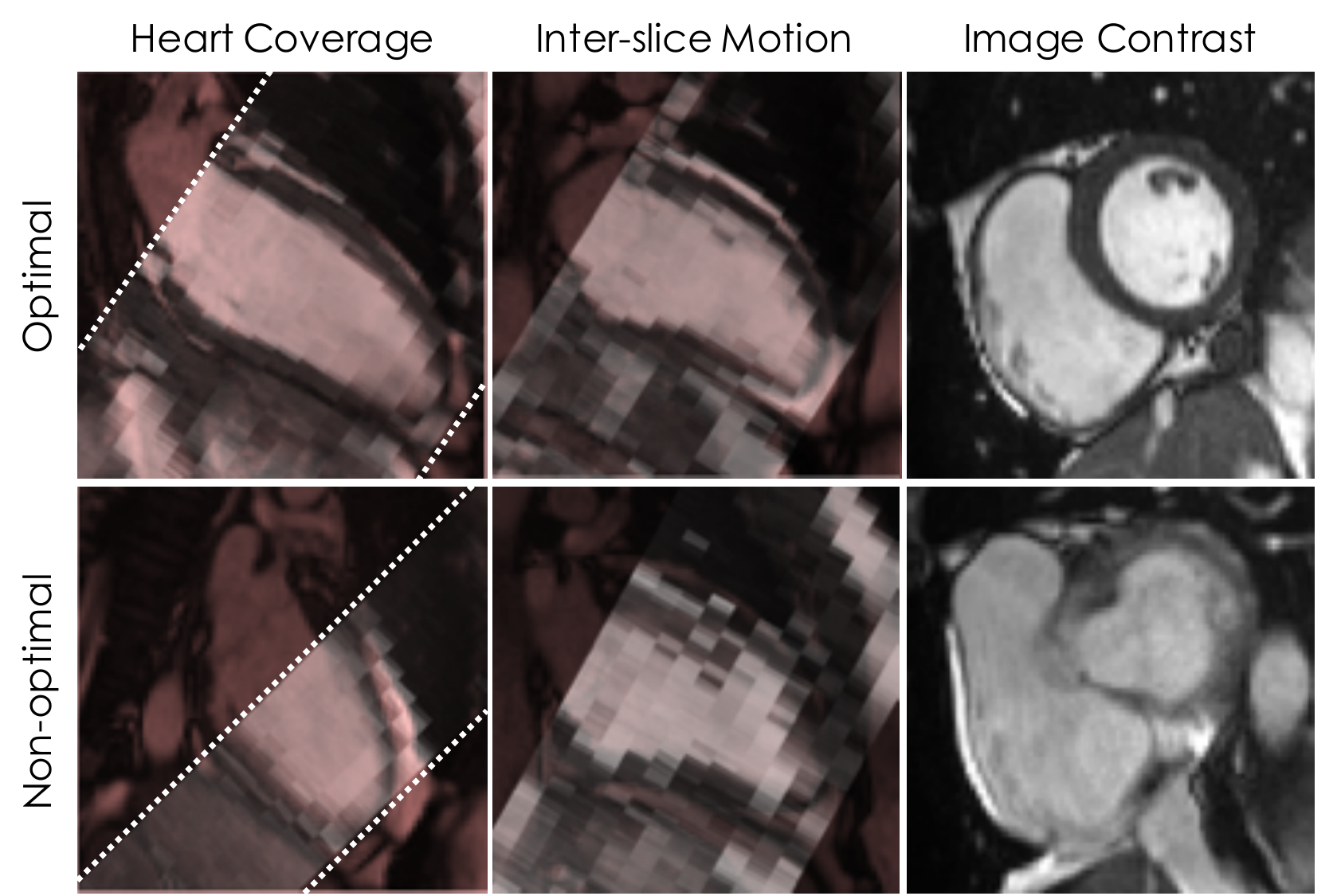}
\caption{Potential issues affecting CMR image acquisitions. In the first two columns, the superimposition of long-axis two-chamber views (red) and short-axis stacks (gray) is shown, while, in the last one, short-axis slices are displayed.}
\label{fig:issues}
\end{figure}
In the last decades, several initiatives for the acquisition of open access large-scale population studies have been launched. For example, the UK Biobank (UKBB) is a population-based prospective study, established to allow detailed investigations of the genetic and non-genetic determinants of the diseases of middle and old age. Of the 500,000 subjects enrolled in the study, CMR will be collected from 100,000 of them \cite{Petersen2016}. At the time of submission the acquisition is ongoing, with close to 20,000 subjects already scanned. Together with this trend towards the implementation of large-scale multi-centre imaging datasets, the need for fast and reliable quality control techniques for CMR images has become evident, as highlighted also by several studies aiming to define standardized criteria for this task \cite{Klinke2013}. In this scenario, quality control through visual inspection is not only subjective, but simply infeasible due to the very high throughput demanded by the acquisition pipeline. On the other hand, failure to correctly identify corrupted or unusable images could affect the results of automated analysis performed on the dataset, with undesirable effects. Consequently, the need for fully automated quality control pipelines for CMR images has arisen.


Many research efforts have been dedicated to the automated identification of quality metrics from MR images. Most of these efforts have focused on the automated estimation of noise levels \cite{Coupe2010,Maximov2012}. Still, many aspects related to the usability of the acquired images are inherently modality-specific. Several automated pipelines for quality control have been proposed for brain MR imaging \cite{Gedamu2008}. However, to our knowledge, no comprehensive automated quality control pipelines have been proposed so far for CMR images, in particular for the short-axis (SA) cine image stacks, which are the reference images for the structural and functional assessment of the heart. One crucial aspect of the acquisition of SA image stacks is that it requires the MR operator to identify the direction of the left ventricular (LV) long axis - the line going from the apex to the centre of the mitral valve - and to define a region of interest: the correct planning will generate a SA stack encompassing both those landmarks with slices perpendicular to the LV long axis. If this selection is incorrect, the acquired SA stack may include an insufficient number of SA slices to fully cover the LV (see first column of Fig. \ref{fig:issues}). As a consequence, any functional analysis performed on the stack (e.g. ventricular volumes estimation) may be compromised. Another important aspect involved in CMR acquisitions is that SA cine stacks are generated during multiple breath-holds (with usually 1-3 slices acquired per each breath-hold). Although the subjects are instructed to hold their breath at the same breath-holding position, in practice the heart location can vary considerably. If the differences between the breath-holding positions are too pronounced, the acquired image stack will be affected by inter-slice motion and \textclr{\blfootnote{R3.4}thus will not correctly represent the cardiac shape}, introducing potential errors in the following analyses and visualizations (see second column of Fig. \ref{fig:issues}). Finally, the contrast of the obtained CMR images is directly affected by the chosen acquisition parameters (as well as by potential artefacts). If the different structures of the heart are not properly contrasted, the assessment of the cardiac function can be hampered (see third column of Fig. \ref{fig:issues}).

In this paper, we present a \textclr{\blfootnote{R3.7}fully-automated, learning-based quality control technique for CMR SA image stacks}. Our approach uses a hybrid decision forest method to extract at once both landmark positions (LMs) and probabilistic segmentation maps (PSMs) from long-axis (LA) and SA images. LMs and PSMs are then used to perform three quality checks: 1) heart coverage estimation, 2) inter-slice motion detection, 3) image contrast estimation in the cardiac region. \textclr{\blfootnote{AE.1,\\R1.2} Our hybrid forest method is thus not intended as a novel technique for landmark detection and segmentation per se, but rather as an integral component of our pipeline. The extraction of LMs from multiple LA views and the probabilistic nature of PSMs allow to assess the reliability of the pipeline for each scan using dedicated sanity checks.} The technique was tested on \textclr{\blfootnote{AE.4,\\R2.8}two datasets (up to 3000 cases from the UKBB study and 100 cases from the UK Digital Heart Project \cite{UKDHP}, UKDHP)} and validated against manual annotations and visual inspections.
\begin{figure*}[!t]
\centering
\includegraphics[width=0.7\textwidth]{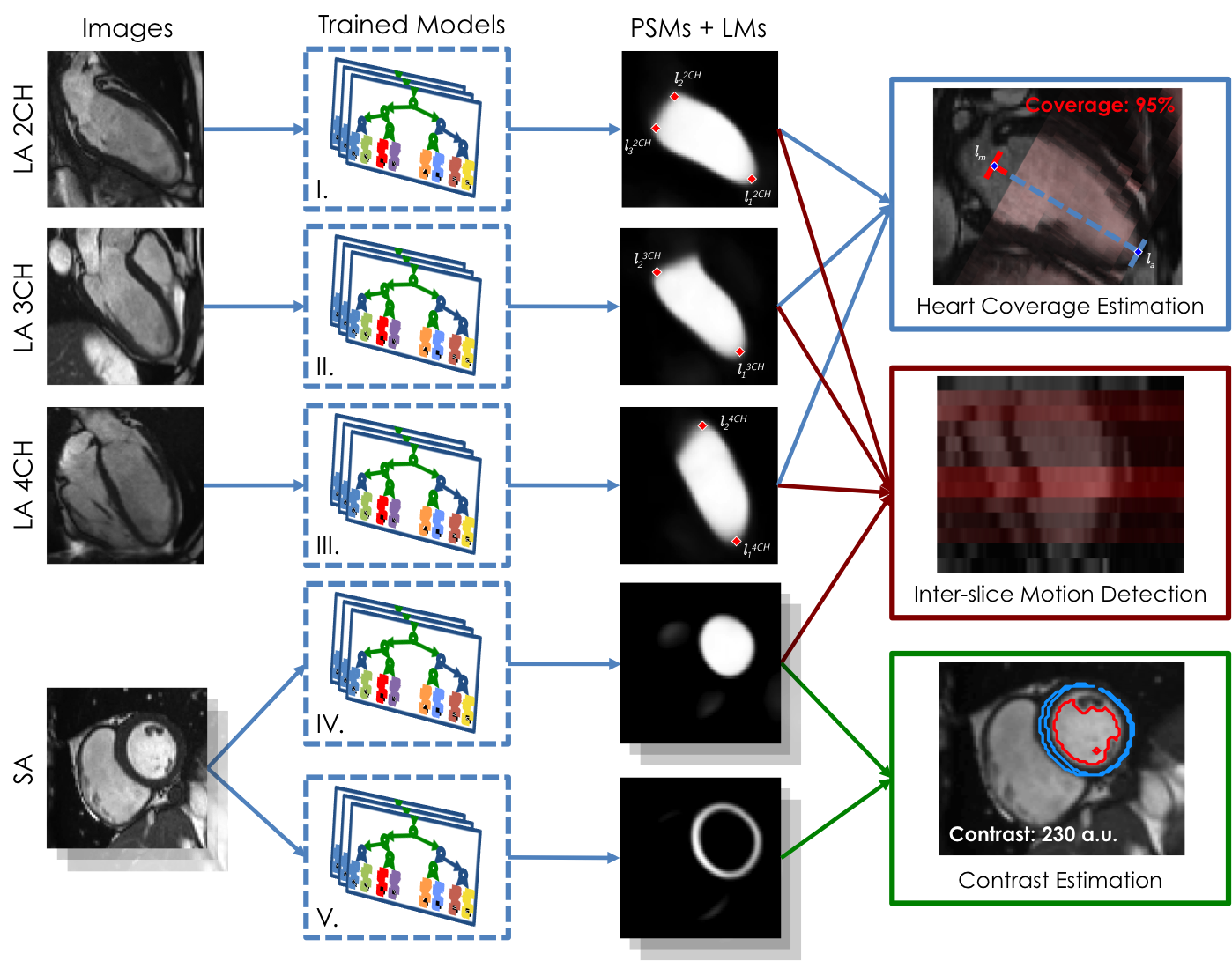}
\caption{Overview of the proposed pipeline. Probabilistic segmentation maps (PSMs) and landmark positions (LMs) are extracted from LA and SA images using hybrid random forests and exploited to perform three separate quality control checks.}
\label{fig:pipeline}
\end{figure*}

\section{Related Work}

To the best of our knowledge, differently from brain MRI \cite{Gedamu2008}, no comprehensive quality control techniques for cardiac CMR images have been reported in the literature. One of the few studies in this direction has been recently presented by Alba et al. \cite{Alba2018}, who however focussed on assessing segmentation quality rather than image quality. On the other hand, automated heart coverage estimation alone has been the aim of several studies. Zhang et al. \cite{Zhang2016, Zhang2017} proposed to use convolutional neural networks (CNN) to perform slice classification in order to detect the presence or absence of the basal and apical slices. In their first work \cite{Zhang2016} they proposed a 2D CNN trained on UKBB data, while in their more recent one \cite{Zhang2017} they improved their previous results by using a generative adversarial network. Differently from these techniques, our approach to heart coverage estimation is based on the detection of landmarks: in our previous preliminary work \cite{Tarroni2017a}, we proposed a decision forest method to detect the cardiac apex and the mitral valve on long-axis 2-chamber (LA 2CH) view images, and used the position of these landmarks with respect to the space encompassed by the acquired stack to estimate the coverage. The technique was applied to 3000 cases extracted from the UKBB, and was able to detect SA stacks with insufficient coverage with relatively high accuracy. 

\textclr{\blfootnote{R3.8}Motion detection and modeling in the thoracic area has been a highly investigated subject for more than a decade \cite{McClelland2013}. As far as inter-slice respiratory motion in CMR is concerned, most of the approaches reported in the last decade have focussed on motion correction rather than motion detection \cite{Lotjonen2004a, Oktay2016, Sinclair2017, Yang2017}}. All of the cited studies focused on the compensation of inter-slice motion and in the generation of a corrected SA stack by means of rigid in-plane registration. Unfortunately, however, respiration causes a complex roto-translation of the heart in all three dimensions \cite{McLeish2002}: while most translation happens in the cranio-caudal direction (thus approximately almost perpendicularly to the long axis of the LV), big differences in subsequent breath-holding positions can cause out-of-plane motion, which would lead to an inaccurate representation of the heart in the stack. As a consequence, it is important to estimate the amount of motion occurred during the acquisition of the stack in order to decide whether there are the grounds for the application of a motion correction technique or it is instead advisable to repeat the scan (or exclude it from subsequent analyses).

In the past, several research efforts have been made towards the correct quantification of signal-to-noise (SNR) or contrast-to-noise (CNR) ratios in MR images \cite{Coupe2010}. However, modern acquisition techniques making use of parallel imaging produce images with spatially-varying noise distributions, rendering image-based estimators unreliable \cite{Dietrich2007}. To overcome this limitation, more elaborate methods have been proposed exploiting information about coil sensitivity or reconstruction coefficients \cite{Aja-Fernandez2014}. Unfortunately, these data are very often not available, making the estimation of noise, and consequently of SNR and CNR, practically unfeasible in most scenarios. At the same time, image contrast between two objects - simply defined as the difference between their signal intensity - has long been used to determine their visual differentiability in the acquired MR image \cite{Wolff1997}. In CMR imaging, images with poor contrast between the LV cavity and myocardium can potentially hinder the assessment of cardiac structure and function: consequently, contrast estimation in the cardiac region can provide a useful metric for quality control purposes, either triggering the use of contrast-enhancing techniques or a new acquisition.

In this paper, we present a fully-automated, learning-based quality control pipeline for CMR SA stacks. \textclr{\blfootnote{AE.3,\\R4.1}The proposed approach builds upon our previous work \cite{Tarroni2017a}, which used a hybrid decision forest method \cite{Oktay2017} to extract LMs from LA 2CH view images in order to perform heart coverage estimation. With respect to our previous approach as well as to state-of-the-art techniques, the main contributions of the present work can be listed as follows:}

\begin{itemize}
\item \textclr{We present the first comprehensive, fast, fully-automated quality control pipeline specifically designed for CMR SA image stacks. The checks incorporated in the pipeline are 1) heart coverage estimation, 2) inter-slice motion detection, 3) image contrast estimation in the cardiac region. To the best of our knowledge, motion detection and cardiac image contrast \blfootnote{R3.9}for the sake of quality control have not been investigated before. As for heart coverage estimation, we build on our previously published study \cite{Tarroni2017a} by extending LMs extraction to all long-axis views. LMs are then combined together to substantially increase the robustness and the reliability of this quality check (for details please refer to the Discussion section);}
\item \textclr{We propose a different implementation of the previously published hybrid decision forest \cite{Oktay2017} (adopted in our previous work \cite{Tarroni2017a}) which allowed the joint extraction of LMs and probabilistic edge maps (PEMs). The new implementation (based on a novel mapping) allows instead the extraction of LMs and PSMs: \blfootnote{R1.3}PSMs are required to perform both inter-slice motion detection and cardiac image contrast estimation, and enable sanity checks to assess the reliability of the pipeline;}
\item We validate this pipeline by applying it to up to 3000 cases extracted from the UKBB study \textclr{\blfootnote{AE.4,\\R2.8}and to 100 cases from the UKDHP}, showing its accuracy and robustness in real world scenarios. The pipeline could be both applied retrospectively on large-scale datasets to improve the reliability of clinical studies or deployed prospectively at acquisition sites to allow almost real-time assessment of the acquired scans.
\end{itemize}

\section{Methods}

The proposed quality control pipeline is summarized in Fig. \ref{fig:pipeline}. All of the three quality control steps are based on the information extracted by hybrid decision forest models from the acquired images. This section of the paper starts with a brief recap on the theory behind decision forests and is followed by the description of the implementation adopted in the proposed pipeline, which allows the joint extraction of LMs and PSMs. Finally, each specific quality control step is described in detail.
\begin{figure*}[!t]
\centering
\includegraphics[width=0.9\textwidth]{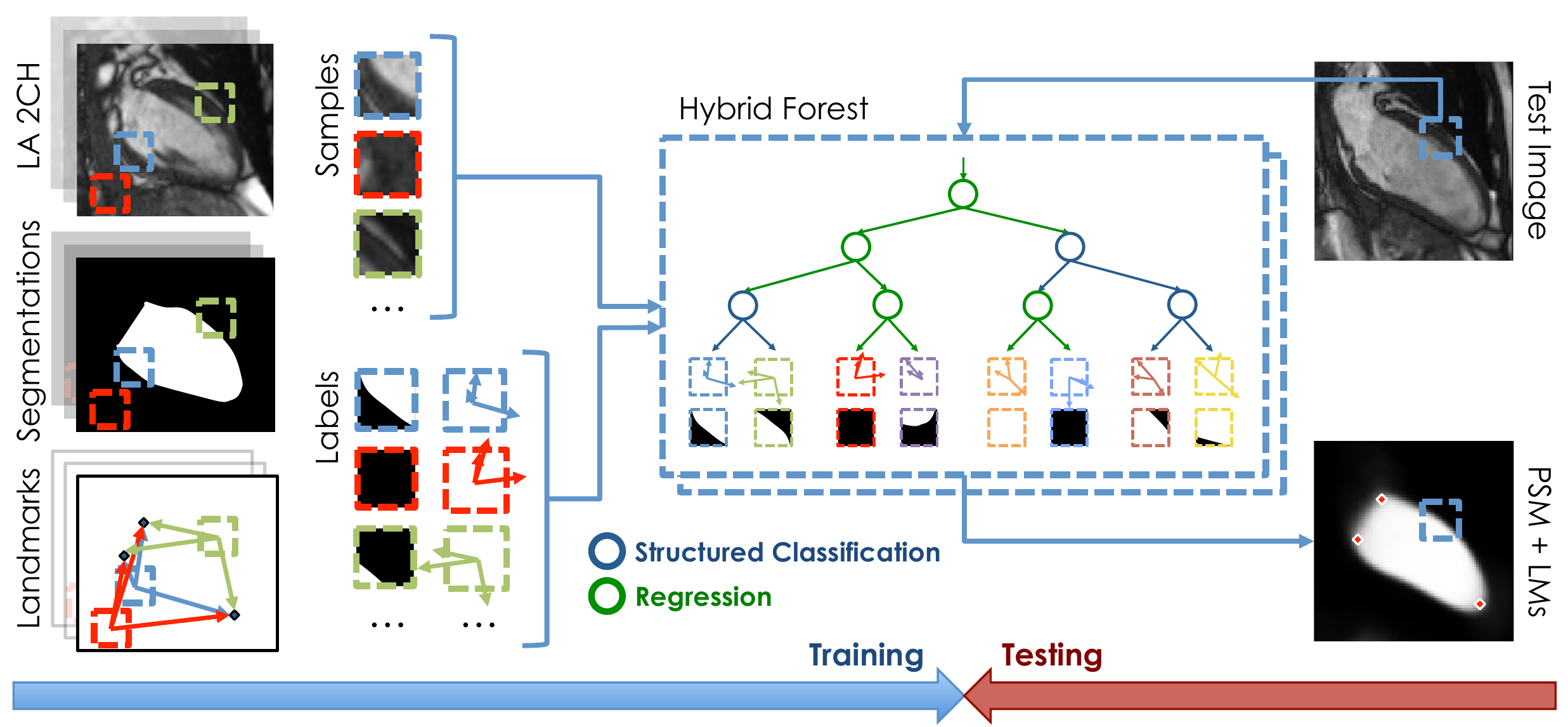}
\caption{Hybrid random forest. During training, randomly extracted samples with associated labels - consisting of segmentations and vector displacements - are fed to the forest, and the learnt associations are stored in the leaf nodes. During testing, each sample extracted from the test image is sent to the model, extracting both PSM and LMs at once.}
\label{fig:forest}
\end{figure*}

\subsection{Hybrid Decision Forests}
A decision tree consists of a combination of split and leaf nodes arranged in a binary tree structure \cite{Criminisi2011}. Trees route a sample $\bm{x} \in \mathcal{X}$ (in our case an image patch) by recursively branching left or right at each split node $j$ until a leaf node $k$ is reached, where the posterior distribution $p(y|\bm{x})$ for the output variable $y \in \mathcal{Y}$ is stored. Each split node $j$ is associated with a binary split function $h(\bm{x},\bm{\theta}_j) \in \{0,1\}$ defined by the set of parameters $\bm{\theta}_j$: if $h = 0$ the node sends $\bm{x}$ to the left, otherwise to the right. Usually $h$ is a decision stump, i.e. a single feature dimension $n$ of $\bm{x}$ is compared with a threshold $\tau$: $\bm{\theta} = (n,\tau)$ and $h(\bm{x},\bm{\theta}) = [ \bm{x}(n)<\tau ]$. A decision forest is an ensemble of $T$ independent decision trees: during testing, given a sample patch $\bm{x}$, the predictions of the different trees are combined into a single output by means of an ensemble model. During training, at each node the goal is to find the set of parameters $\bm{\theta}_j$ which maximizes a previously defined information gain $I_j$, usually defined as $I_j = H(S_j) - \sum_{i \in \{0,1\}} \nicefrac{|S_j^i|}{|S_j|}\cdot H(S_j^i)$, where $S_j$, $S_j^0$ and $S_j^1$ are respectively the training set (comprising of samples $\bm{x}$ and associated labels $y$) arriving at node $j$, leaving the node to the left and to the right. $H(S)$ is the entropy of the training set, whose construction depends on the task at hand (e.g. classification, regression). Different types of nodes (maximizing different information gains) can be interleaved within a single tree structure (hence named ``hybrid") in order to perform multiple tasks. As in previous approaches \cite{Gall2011, Oktay2017}, in the proposed technique structured classification nodes (aiming at the detection of an object close to the desired landmarks, in our case usually the LV cavity) and regression nodes (aiming at landmark localization) are combined (see Fig. \ref{fig:forest}). In particular, in the proposed framework, landmark localization is conditioned on the results of the detection of the cavity \cite{Gall2011}. This not only leads to the extraction of two different types of information (PSMs and LMs) with only one model, but improves landmark localization by implicitly incorporating complementary information about cardiac position and shape.

\subsubsection*{Structured Classification and PSM Extraction} Structured classification extends the concept of classification by using structured labels for $\mathcal{Y}$ instead of integer labels. In our case, each label $\bm{y} \in \mathcal{Y}$ (associated with the image patch $\bm{x}$) consists of a segmentation of the LV cavity within $\bm{x}$. To train a structured classification node it is necessary to find a way to cluster structured labels at each split node into two subgroups depending on a similarity measure. The solution to this problem was first proposed by Dollar et al. \cite{Dollar2015} and consists of two steps. First, $\mathcal{Y}$ is mapped to an intermediate space $\mathcal{Z}$ by means of the function $\Pi: \mathcal{Y} \rightarrow \mathcal{Z}$ where the distance between labels can be computed. Importantly, $\Pi$ must be chosen so that similar labels $\bm{y}$ will be associated with vectors $\bm{z}$ close to each other with respect to the distance defined in $\mathcal{Z}$. Then, PCA is applied to the vectors $\bm{z}$ to map the associated labels $\bm{y}$ into a binary set of labels $c \in \mathcal{C}$ $ = \{0,1\}$, which is achieved by applying a binary quantization to the principal component of each $\bm{z}$ vector. Finally, the Shannon entropy can be adopted \cite{Dollar2015}:
\begin{equation}
H_{sc}(S) = - \sum_{c \in \mathcal{C}} p(c)log\big(p(c)\big),
\label{eq_shannon}
\end{equation}
with $p(c)$ indicating the empirical distribution extracted from the training subset at each node. In our previous work \cite{Oktay2017}, this approach has been adopted for structured labels $\mathcal{Y}$ consisting of edge maps (EMs) highlighting the contours of the myocardium. In the case of EMs, the mapping $\Pi$ can simply encode for each pair of pixels whether they belong to the same segment in the label $\bm{y}$ or not: 
\begin{equation}
\Pi_{EM}: \bm{z} = [\bm{y}(j_1)=\bm{y}(j_2)] \qquad \forall j_1 \neq j_2,
\label{eq_discretize_pem}
\end{equation}
where $j_1$ and $j_2$ are indices spanning every pixel in $\bm{y}$ \cite{Dollar2015}. The resulting long binary vector $\bm{z}$ \textclr{\blfootnote{R4.5}(which has a number of dimensions equal to the number of pixel pairs in $\bm{y}$)} can be used to compare this particular label to the other ones by simply computing the Euclidean distance in $\mathcal{Z}$. However, the same choice for $\Pi$ cannot be adopted for our task, which aims at using structured labels consisting of segmentation maps (SMs) of the LV cavity. For example, let's imagine two labels $\bm{y}_1$ and $\bm{y}_2$, the former completely outside the LV cavity and the latter completely inside: using the mapping $\Pi_{EM}$, we would obtain $\bm{z}_1 = \bm{z}_2$, which contradicts the requirement by which only similar labels will be mapped close to each other in $\mathcal{Z}$. Consequently, we implemented a different mapping:
\begin{equation}
\begin{aligned}
\Pi_{SM}: \bm{z} = & [\bm{y}(j_1)=\bm{y}(j_2)=0] \oplus \dots \\
\dots & [\bm{y}(j_1)=\bm{y}(j_2)=1] \qquad \forall j_1 \neq j_2,
\end{aligned}
\label{eq_discretize_psm}
\end{equation}
which encodes for each pair of pixels in $\bm{y}$ whether they are both equal to 0, whether they are both equal to 1 and then concatenates the two obtained binary vectors. This formulation ensures the proper computation of the distance between labels, and thus their clustering at each node based on their similarity. At the end of the training process, the label $\hat{\bm{y}}$ stored in each leaf node is the one whose $\hat{\bm{z}}$ is the medoid (i.e. that minimizes the sum of distances to all the other $\bm{z}$ at the same node). At testing time, each sample patch of the test image is sent down each tree of the forest, and the segmentation maps stored at each selected leaf node are averaged, producing a smooth segmentation map (PSM) of the LV cavity. The values in the PSM are actual probabilities (proportional to the certainty in LV cavity detection), and can be used to assess the reliability of the prediction. Of note, the introduced formulation for $\Pi_{SM}$ in Eq. \ref{eq_discretize_psm} could be easily extended to multi-label PSM generation by concatenating additional binary vectors computed for each label $c_i$ and by performing a channel-based averaging operation at testing time.

\subsubsection*{Regression and Landmark Detection} To train regression nodes, it is necessary to associate with each sample patch $\bm{x}$ an additional label $\mathcal{D} = (\bm{d}^1, \bm{d}^2, \dotsc, \bm{d}^L)$, where $\bm{d}^l$ represents for each of the $L$ landmarks the $N$-dimensional displacement vector from the patch centre to the landmark location. Instead of the Shannon entropy defined in Eq. \ref{eq_shannon}, regression nodes are trained by minimizing the determinant of the covariance matrix $|\Lambda(S)|$ defined by the landmark displacement vectors:
\begin{equation}
H_{r}(S) = \frac{1}{2} log\big((2\pi e)^d|\Lambda(S)|\big).
\label{eq_regr}
\end{equation}

Landmark positions are assumed to be uncorrelated, thus only the diagonal elements of $\Lambda(S)$ are used in Eq. \ref{eq_regr} \cite{Criminisi2013}. The location predictions are stored at each leaf node $k$ using a parametric model following a $N\cdot L$-dimensional multivariate normal distribution with $\overline{\bm{d}_k^l}$ and $\Sigma_k^l$ mean and covariance matrices, respectively. At testing time, Hough vote maps are generated for each landmark by summing up the posterior distributions obtained from each tree for each patch \textclr{\blfootnote{R4.10}(applying normalization factors)} \cite{Gall2011}. Assuming that pixels belonging to the LV are more informative for cardiac landmark detection than background ones, the PSM values for the LV cavity are used for each patch as weighting factor during the generation of the $L$ Hough vote maps, effectively restricting voting rights only to pixels likely to belong to the LV cavity itself \cite{Oktay2017}. Finally, the location of a landmark is determined by identifying the pixel with the highest value on each Hough vote map.

\subsubsection*{Model Training} Each patch $\bm{x}$ is represented by several features: multi-resolution image intensity, histogram of gradients (HoG) and gradient magnitude. For a detailed description, please refer to \cite{Oktay2017}. The described hybrid random forest approach is used to build five different models (I-V) for our application (see Fig. \ref{fig:pipeline}): PSM estimation of LV cavity and LMs extraction for apex and mitral valve for LA images, PSM of LV cavity and LV myocardium for SA stacks. For LA 3CH and 4CH images (models II and III) only one mitral valve point is identified because in these images the LV outflow tract of the aorta can partially occlude one side of the mitral valve, making its localization inaccurate. Also, the training of the models using SA images (models IV and V) is performed by feeding the random forests with all the slices extracted from the SA image stacks: consequently, at testing time, the models are applied to each slice of the stack independently.
\begin{algorithm}
\DontPrintSemicolon
\textbf{Input landmarks:} \;
\qquad Apex: $\bm{l}^{2 CH}_1$, $\bm{l}^{3 CH}_1$, $\bm{l}^{4 CH}_1$ \;
\qquad Mitral Valve points: $\bm{l}^{2 CH}_2$, $\bm{l}^{2 CH}_3$, $\bm{l}^{3 CH}_2$, $\bm{l}^{4 CH}_2$ \;
\textbf{Change coordinate system:} \;
\qquad Apex: $\hat{\bm{l}}^{2 CH}_1$, $\hat{\bm{l}}^{3 CH}_1$, $\hat{\bm{l}}^{4 CH}_1$ \;
\qquad Mitral Valve points: $\hat{\bm{l}}^{2 CH}_2$, $\hat{\bm{l}}^{2 CH}_3$, $\hat{\bm{l}}^{3 CH}_2$, $\hat{\bm{l}}^{4 CH}_2$ \:
\textbf{Compute median landmarks:} \;
\qquad $\bm{l}_a = \textrm{median}\big(\hat{\bm{l}}^{2 CH}_1, \hat{\bm{l}}^{3 CH}_1, \hat{\bm{l}}^{4 CH}_1\big)$ \;
\qquad $\bm{l}_m = \textrm{median}\big(\hat{\bm{l}}^{2 CH}_2, \hat{\bm{l}}^{2 CH}_3, \hat{\bm{l}}^{3 CH}_2, \hat{\bm{l}}^{4 CH}_2\big)$ \;
\qquad with z-components $l_a$ and $l_m$, respectively \;
\textbf{Extract SA stack extension in the z direction:} \;
\qquad Apex: $r_a$ \;
\qquad Base: $r_m$ \;
\textbf{Compute coverage CV:} \;
\qquad $ CV =
    \begin{cases}
      \frac{max \big(0,min(r_a,l_a)- max(r_m,l_m)\big)}{l_a-l_m} & \text{if (condition)}\\
      \frac{r_a-r_m}{l_a-l_m} & \text{otherwise}
    \end{cases}      
$ \; \;
\qquad (condition): \text{$r_a<l_a$ or $r_m>l_m$} \;
    \caption{Heart Coverage Estimation}
    \label{algo:coverage}
\end{algorithm}

\subsection{Heart Coverage Estimation}
Heart coverage is estimated exploiting the landmarks identified on LA 2CH, 3CH and 4CH images using the previously trained hybrid forest models. The rationale is that a properly scanned SA stack should encompass the whole portion of space between the apex and the mitral valve. As highlighted in Fig. \ref{fig:pipeline}, for a specific subject we identify three landmarks for the apex (one per each LA image: $\bm{l}^{2 CH}_1$, $\bm{l}^{3 CH}_1$ and $\bm{l}^{4 CH}_1$) and four for the mitral valve ($\bm{l}^{2 CH}_2$, $\bm{l}^{2 CH}_3$, $\bm{l}^{3 CH}_2$, $\bm{l}^{4 CH}_2$) with values in the coordinate systems of each respective LA image. Using the orientation matrix extracted from the DICOM headers of the acquired SA and LA images, it is possible to define the coordinates of these landmarks in the coordinate system of the SA stack itself. Two new ``median" landmarks ($\bm{l}_a$ and $\bm{l}_m$) are then defined taking the medians of the coordinates of the landmarks for the apex and for the mitral valve, respectively, in the SA coordinate system. The extension in the z direction (i.e. along the LV long axis) of the SA stack can be easily computed from the slice thickness and slice number, which are stored in the DICOM header of the stack itself: the two extrema along this direction are defined $r_a$ and $r_m$, respectively. Finally, the relative coverage can be computed by comparing the relative positions along the z direction of $\bm{l}_a$ and $\bm{l}_m$ (i.e. the space that is supposed to be covered by the SA stack) to the portion of space between $r_a$ and $r_m$ (i.e. the space that is actually covered). The steps for coverage estimation are listed in Algorithm \ref{algo:coverage}, including the formula for the computation of the coverage (under the assumption that the apex is located at higher z compared to the mitral valve). Importantly, this technique can seamlessly be applied even if only one LA image is available. Also, while minor motion can occur between the acquisitions of LA images and of the SA stack, it is generally negligible in the z direction (the only one influencing coverage) \cite{McLeish2002} and thus registration procedures between these images were found to be unnecessary. \textclr{\blfootnote{R1.5}Finally, a sanity check is performed to detect cases in which landmark detection failed: for each LA view, when either of the relative distances between the landmarks was greater or smaller than reference values by a certain threshold, the landmarks from that image were discarded, and the automated coverage estimation was performed only on the remaining landmarks (if available)}. 

\subsection{Inter-Slice Motion Detection}
\begin{figure}[!t]
\centering
\includegraphics[width=\columnwidth]{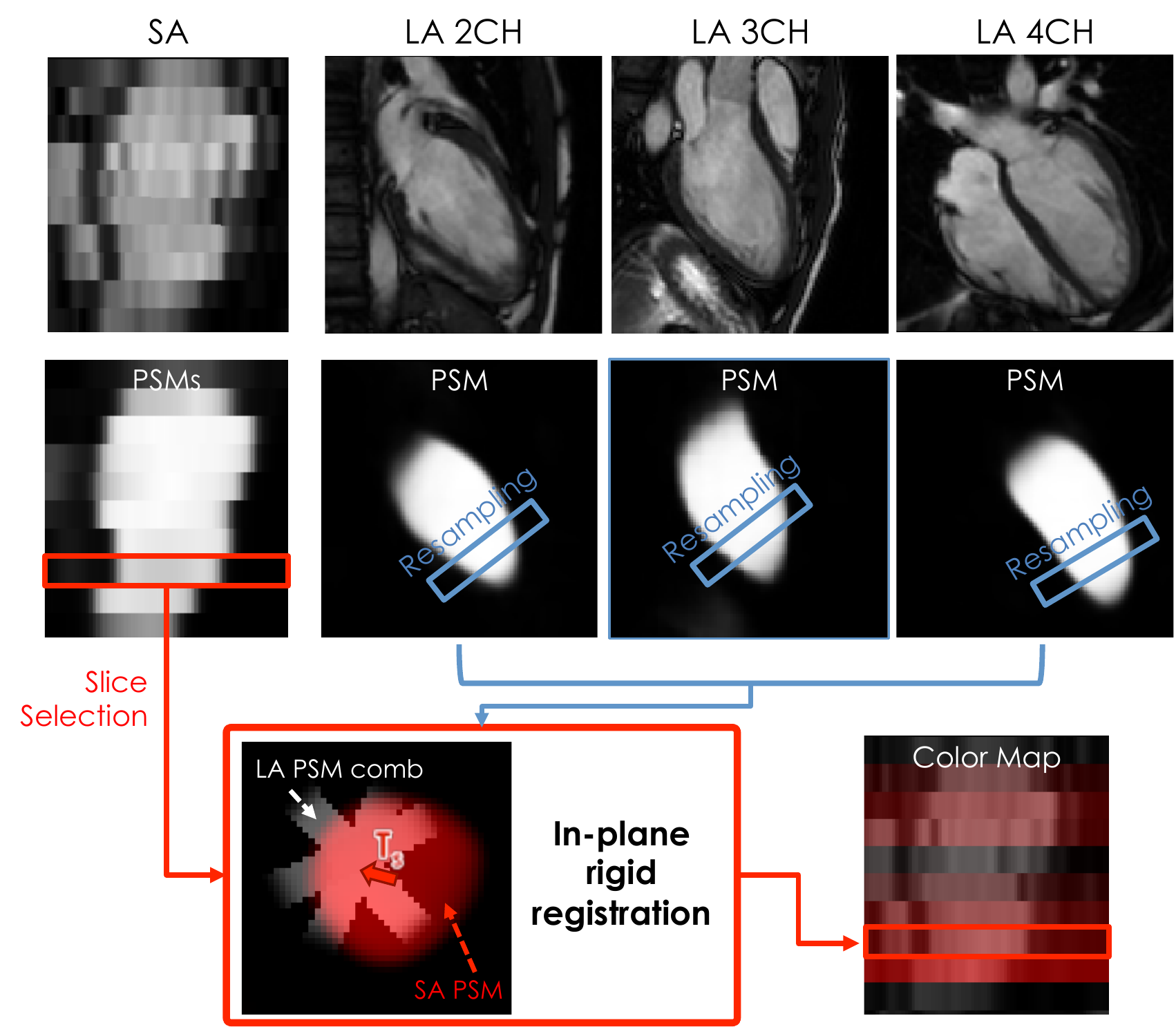}
\caption{Motion detection technique. For each slice of the SA stack, the corresponding portion of space in each LA PSM is resampled and combined, producing the ``asterisk-shaped" LA PSM comb image. In-plane rigid registration is then performed between each SA PSM and LA PSM comb, and the translation magnitude $T_s$ used as proxy for inter-slice motion for that slice. A color map, with the intensity of each slice proportional to the respective $T_s$, can be also generated.}
\label{fig:motion:pipeline}
\end{figure}

Inter-slice motion detection relies on the PSMs extracted from the acquired images. The rationale is that while LV cavity PSMs of motion-corrupted SA slices are misaligned, PSMs extracted from the LA images represent sections of the true shape of the LV cavity and can consequently be used as reference. Moreover, the amount of misalignment between the SA PSMs and LA PSMs can be used as an indicator of motion. To perform this assessment, the LA PSMs are initially rigidly registered (by 3D translation only, using sum of squared differences as dissimilarity metric) to the SA PSM stack to compensate for potential motion between different acquisitions. Then, for each slice of the SA PSM stack, the three registered LA PSMs are resampled and combined into a single image (referred to as combined LA PSM) containing the sections of the LA PSMs with respect to a specific slice (see Fig. \ref{fig:motion:pipeline}). Finally, in-plane rigid registration (by translation only, using sum of squared differences as dissimilarity metric) is performed between each SA PSM slice and the associated combined LA PSM, and the magnitude of the translation $T_s$ used as a metric for motion (i.e. differences in breath-holding positions). Of note, this step is performed only on the slices which are effectively covering the LV, condition assessed using the LA LMs as in Algorithm \ref{algo:coverage}. \textclr{\blfootnote{AE.1,\\R1.2}The probabilistic nature of PSMs allows the application of a sanity check performed to detect slices with a failed PSM estimation: SA PSM slices (whose values range between 0 and 1024) with a peak probability value below a user-defined threshold are considered unreliable, and thus their $T_s$ discarded.} \textclr{Also, this technique could be applied even if only two LA images were available.} The steps for motion detection are listed in Algorithm \ref{algo:motion}. 

\begin{algorithm}
\DontPrintSemicolon
\textbf{Input PSMs:} \;
\qquad LA images: $PSM^{2CH}$, $PSM^{3CH}$, $PSM^{4CH}$ \;
\qquad SA slices: $PSM^{SA-Cav}_{s}$, $s = (1, \dots, numSlices)$ \;
\textbf{Perform rigid registration of LA PSMs to SA PSM:} \;
\qquad Output: $\overline{PSM}^{2CH}$, $\overline{PSM}^{3CH}$, $\overline{PSM}^{4CH}$ \;
\For{$s = 1$ \KwTo numSlices}{
    \textbf{Resample LA PSMs:} \;
    \qquad Output: $\overline{PSM}^{2CH}_s$, $\overline{PSM}^{3CH}_s$, $\overline{PSM}^{4CH}_s$ \;
    \textbf{Combine resampled LA PSMs:} \;
    \qquad Output: $\overline{PSM}^{LA\_comb}_s$ \;
    \textbf{Perform in-plane rigid registration of} $PSM^{SA-Cav}_{s}$ \textbf{to} $\overline{PSM}^{LA\_comb}_s$\textbf{:} \;
    \qquad Output: Translation magnitude $T_s$ \;
    }
    \caption{Inter-Slice Motion Detection}
    \label{algo:motion}
\end{algorithm}

\subsection{Cardiac Image Contrast Estimation}
Cardiac image contrast is estimated using the LV cavity and LV myocardium PSMs extracted from the SA stack. The rationale is to transform the PSMs into hard segmentations (SMs) and to use them to estimate the difference between average pixel intensity in the LV cavity and in the LV myocardium. Each cavity PSM slice is thresholded selecting the $N_{cav}$ pixels with the highest probability values: this will maximize the probability of measuring the intensity in the actual cavity. The same happens to each myocardium PSM, thresholded selecting $N_{myo}$ pixels. To exclude potential spurious regions from the obtained segmentation, the average centroid for the cavity segmentation is computed among the different slices, and for each slice only the connected component closest to the average centroid is kept, both for the cavity and the myocardium segmentations. Of note, this step is performed taking into account the slice-by-slice rigid transformation estimated using Algorithm \ref{algo:motion}, which amounts to performing the average centroid computation and connected components analysis on a motion-compensated stack. Finally, in order to exclude potential papillary muscles from the cavity intensity computation, a Gaussian mixture model is fitted to the distribution of intensity values inside the cavity segmentation. Since only some slices show papillary muscles, both a two-component and a one-component models are used, and only one is selected based on the Akaike information criterion \cite{Celeux1996}. If the two-component model yields the best fit, since the cavity distribution is always higher than that of papillary muscles, the mean of the component with the highest mean is used as average intensity value for the cavity. For the myocardium, the mean intensity of the pixels masked by the segmentation is computed. Cardiac image contrast is finally defined as the difference between these two values. \textclr{\blfootnote{AE.1,\\R1.2}A double sanity check is performed leveraging the probabilistic nature of PSMs: if either the peak value of either the cavity or the myocardium PSM was below a user-defined threshold or the size of either of the final hard segmentations for the cavity or the myocardium was less than a defined number of pixels, the obtained contrast was deemed unreliable}. The steps for cardiac image contrast estimation are also listed in Algorithm \ref{algo:contrast}.

\begin{algorithm}
\DontPrintSemicolon
\textbf{Input PSMs:} \;
\qquad LV cavity: $PSM^{SA-Cav}_{s}$ \;
\qquad LV myocardium: $PSM^{SA-Myo}_{s}$ \;
\qquad with $s = (1, \dots, numSlices)$ \;
\For{$s = 1$ \KwTo numSlices}{
    \textbf{Threshold} $PSM^{SA-Cav}_{s}$ and $PSM^{SA-Myo}_{s}$\textbf{:} \;
    \qquad Output: $SM^{SA-Cav}_{s}$ and $SM^{SA-Myo}_{s}$ \;
    \textbf{Estimate centroids $\mathbf{P}_s$ for $SM^{SA-Cav}_{s}$} \;
}
\textbf{Estimate mean centroid:} $\mathbf{P} = $ mean($\mathbf{P}_s$) \;
\For{$s = 1$ \KwTo numSlices}{
    \textbf{Exclude all but one connected component per SM based on distance to $\mathbf{P}$:} \;
    \qquad Output: $\overline{SM}^{SA-Cav}_{s}$ and $\overline{SM}^{SA-Myo}_{s}$ \;
    \textbf{Fit Gaussian Mixture Model to} $\overline{SM}^{SA-Cav}_{s}$ \textbf{to exclude papillary muscles:} \;
    \qquad Output: $\mu^{SA-BP}_{s}$ \;
    \textbf{Compute contrast CT:} \;
    \qquad $CT = \mu^{SA-BP}_{s}$ - mean$\Big(SA_s\Big(\overline{SM}^{SA-Myo}_{s}\Big)\Big)$\;
    \qquad with $SA_s$ the s-slice of the SA stack \;
    }
    \caption{Cardiac Image Contrast Estimation}
    \label{algo:contrast}
\end{algorithm}

\subsection{Performance Evaluation}

\subsubsection*{Image Acquisition}
\textclr{\blfootnote{AE.4}To train and test the proposed quality control pipeline, images from two different datasets were used: the UKBB \cite{Petersen2016} and the UKDHP \cite{UKDHP}.} CMR imaging for the UKBB was performed using a 1.5T Siemens MAGNETOM Aera system equipped with a 18 channels anterior body surface coil (45 mT/m and 200 T/m/s gradient system). 2D cine balanced steady-state free precession (b-SSFP) SA image stacks were acquired with in-plane spatial resolution 1.8$\times$1.8 mm, slice thickness 8 mm, slice gap 2 mm, image size 198$\times$208 and average number of slices 10. 2D cine b-SSFP LA images were acquired with in-plane spatial resolution 1.8$\times$1.8 mm, slice thickness 8 mm and image size 162$\times$208. Further acquisition details can be found in \cite{Petersen2016}.
\textclr{\blfootnote{AE.4,\\R2.8}CMR imaging for the UKDHP was performed on healthy volunteers using a 1.5T Philips Achieva system equipped with a 32 element cardiac phased-array coil (33 mT/m and 160 T/m/s gradient system). 2D cine balanced steady-state free precession (b-SSFP) SA image stacks were acquired with in-plane spatial resolution 1.2$\times$1.2 mm, slice thickness 8 mm, slice gap 2 mm, image size 288$\times$288 and average number of slices 12. 2D cine b-SSFP LA images were acquired with in-plane spatial resolution 1.5$\times$1.5 mm, slice thickness 8 mm and image size 256$\times$256.}
In both datasets, only end-diastolic frames were considered.

\subsubsection*{Experimental design}
A series of experiments was conducted to assess the accuracy of each portion of the pipeline. First of all, the five hybrid random forest models were trained using a randomly-generated subset of 500 cases from the UKBB. For each LA image-based model, the 500 images were used together with manually-annotated landmarks and segmentations of the LV cavity. \textclr{\blfootnote{R1.4}The segmentations were obtained with a CNN-based automated tool proven to reach human-level performance \cite{Bai2017a}, and then visually checked for accuracy.} Each training set was quadrupled in size through data augmentation applying random rescaling (following a normal distribution with $\mu = 1$, $\sigma = 0.1$) and random rotation ($\mu = 0\degree$, $\sigma = 30\degree$). For each of the two SA stack-based models, the slices extracted from the 500 stacks were used (for a total of 5165 images) together with segmentations of the LV cavity and of the LV myocardium, respectively \textclr{\blfootnote{R1.4}(obtained using the same process described for LA images)}. Details regarding forest training include image patch size 48$\times$48 px for LA models and 32$\times$32 px for SA ones, segmentation label size 16$\times$16 px, number of samples 4$\cdot10^6$, number of trees $T =$ 8. 

A first series of experiments was performed by evaluating the trained pipeline on a separate testing set consisting of 3000 cases randomly extracted from the UKBB. To evaluate the accuracy of the proposed heart coverage estimation technique, two experiments were conducted. First, for each of the three LA views, the positions of the landmarks were manually annotated on 100 randomly selected cases. The automatically detected LMs were compared to the manually identified ones by measuring the Euclidean distance between the two sets of points. Then, the 3000 SA stacks were visually inspected (sometimes using LA images as reference) to identify cases with insufficient coverage, defined as such when at least one full slice was missing. Automated heart coverage estimation was then performed on the same dataset. \textclr{\blfootnote{R1.5,\\R3.15}To instruct the previously described sanity check, the mean and standard deviation of the relative distances between manually annotated landmarks were computed on the 100 images ($\overline{\bm{l}_2-\bm{l}_1} = 89 \pm 12$ mm, $\overline{\bm{l}_3-\bm{l}_2} = 32 \pm 5$ mm, $\overline{\bm{l}_3-\bm{l}_1} = 87 \pm 12$ mm): then, for each LA view, when either of the relative distances between the automatically detected LMs was over 2 standard deviations greater or smaller than the respective mean distance value (thus covering roughly 95\% of the measured variability), the LMs from that image were discarded and the automated coverage estimation was performed only on the remaining ones (if available).} Finally, the accuracy of the technique was assessed against the performed visual inspection \textclr{\blfootnote{AE.4}performing a standard binary classification test using a threshold for insufficient coverage optimized automatically with an ROC analysis.}
To evaluate the accuracy of the motion detection technique, \textclr{\blfootnote{R4.7,\\R2.8}two experiments were conducted. First, for each of the three LA views as well as for the SA stacks, the automatically extracted PSMs were compared to hard segmentations obtained using the previously-described CNN-based automated tool \cite{Bai2017a} on 1000 randomly selected cases. While this experiment was aimed at assessing the accuracy of the PSMs, it is worth noting that the PSMs are never directly thresholded for segmentation purposes in the pipeline, which on the contrary exploits their probabilistic nature. For the sake of this comparison, the PSMs were turned into hard segmentation by applying a global threshold and compared to the reference ones by computing the Dice coefficient (DSC). The global threshold was optimized automatically using an ROC analysis.} Then, 1500 SA stacks were visually inspected (sometimes using LA images as reference) to identify cases with noticeable motion corruption. Automated motion detection was then performed on the same dataset. To implement the previously described sanity check, PSM slices with peak probability values below 600 were considered not reliable for motion detection, and thus their $T_s$ (i.e. the estimated translation magnitude) discarded; if less than 2 $T_s$ values were left, the motion detection analysis was not performed on the specific stack. Accuracy of the automated technique was assessed against visual inspection with a standard binary classification test using the following criterion: a stack was deemed motion-corrupted if either the average $T_s$ was above a first threshold $T_A$ or at least two $T_s$ were above a second threshold $T_B$. This double criterion aimed at the detection of both stacks with a few, clearly misaligned slices as well as stacks with poor general alignment. Both $T_A$ and $T_B$ were optimized automatically using an ROC-like approach.
To evaluate the accuracy of the cardiac image contrast estimation technique, 100 random slices from as many random SA stacks were manually annotated selecting regions of interests (ROIs) within the LV cavity and the LV myocardium. Cardiac image contrast was estimated both from the original images and from the images after contrast normalization using a randomly selected reference image stack. Automated contrast estimation was then performed on the same dataset, both before and after normalization, using $N_{cav} = $ 450 px and $N_{myo}= $ 200 px. To implement the previously described sanity check, contrast extracted from slices with PSMs (either for the cavity or for the myocardium) with peak values below 150 or with respective hard segmentations \textclr{\blfootnote{AE.4}with a size of less than 32 $mm^2$ (i.e. 10 pixels)} was deemed unreliable and excluded from the analysis. Automatically estimated and manually computed contrast values were compared using Pearson's correlation coefficient, linear regression and Bland-Altman analyses.
\begin{figure*}[!t]
\centering
\includegraphics[width=0.8\textwidth]{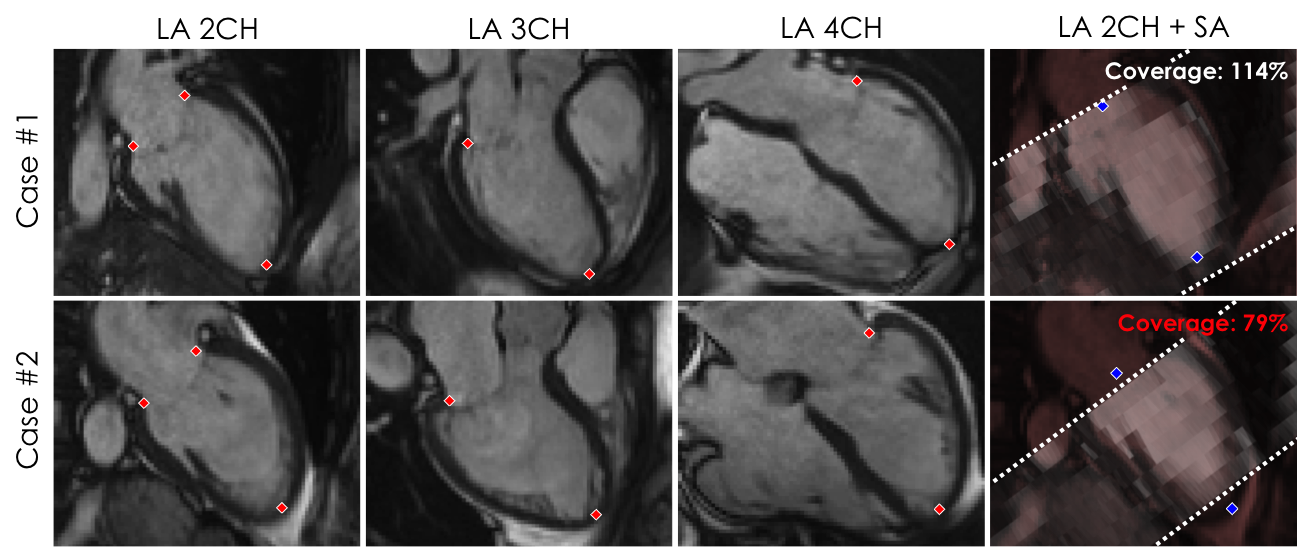}
\caption{Results for heart coverage estimation in two cases, one with sufficient (case \#1) and one with insufficient coverage (case \#2). In the first three columns, the results for landmark detection in the three LA views. In the last column, a mix view with the LA two-chamber view and the SA stack together with the median landmarks for the apex and the mitral valve.}
\label{fig:coverage:results}
\end{figure*}

\textclr{\blfootnote{AE.4,\\R2.8}A second series of experiments was then performed by evaluating the pipeline trained on UKBB on a separate testing set consisting of 100 cases randomly extracted from the UKDHP. Since the scans in UKDHP were acquired with a different scanner and with different parameters from those used for UKBB, these experiments were aimed at assessing the generalization properties of the proposed pipeline. To harmonize the differences between training and testing datasets, the images in UKDHP were pre-processed through intensity normalization \cite{Nyul2000}, spatial resampling and image reorientation. The 100 cases were then visually inspected and manually annotated following the same criteria described for the previous experiments to provide the ground truth for estimation coverage, motion detection and contrast estimation. Since the visual assessment for heart coverage estimation returned no sub-optimal cases, a procedure was implemented to simulate coverage issues and allow a more meaningful evaluation of the pipeline. Stacks were randomly picked following a uniform distribution (10\% chances of being picked), and a number of slices were deleted (either from the top or the bottom of the stack with equal probability), with this number randomly selected from a normal distribution ($\mu = 1$, $\sigma = 2$). Coverage was then visually re-assessed on the whole dataset. It is important to note that while this corruption procedure altered the properties of the dataset with respect to coverage, it did not affect the images on which the learning-based portion of the pipeline is applied (i.e. the LA images) but only the SA stacks, which influence the coverage estimation by means of their size and spatial orientation. The pipeline was applied with the same settings used for the previous dataset except for the threshold for the sanity check for contrast estimation relative to the peak PSM value, which was moved from 150 to 100 to account for the slightly lower overall response in the PSMs. The evaluation strategy for the three checks was the same as for the previous set of experiments.}

\textclr{\blfootnote{AE.2,\\R1.4,\\R1.8,\\R2.9}For all the experiments, manual annotations and visual inspections used as ground truth were performed internally by G. T. (medical imaging researcher with 10 years of experience in cardiac imaging) and H. S. (experienced cardiologist), both blinded to the results of the automated analyses: more specifically, H. S. visually inspected the 3000 SA stacks from the UKBB dataset to identify cases with insufficient coverage, and G. T. performed all of the remaining assessments.}

\section{Results}

The experiments were initially run on a single core of an Intel\texttrademark Xeon CPU E5-1650 v3 @ 3.50GHz with 64 GB of memory to assess the speed of the current pipeline. Average time required to extract PSMs and LMs (when included in the model) was 1.3s per SA stack (of roughly 10 slices) and 0.85s per LA image. Average times required to perform the quality control checks were 0.26s per SA stack for coverage estimation, 9s per SA stack for motion detection (in this case using parallelization on 6 cores to evaluate multiple slices from one stack at once) and 0.6s per slice for contrast estimation.

The localization errors for landmark detection on UKBB for the three LA views are reported in Table \ref{tab:lms} and in Fig. 9 (Supplementary Material). Of note, the landmarks extracted from one image per LA view were identified as outliers and thus excluded from the reported results. \textclr{\blfootnote{R4.7,\\R2.8}Mean DSC values between thresholded PSMs (using a threshold of 450) and reference segmentations were respectively 0.90 $\pm$ 0.07 for the SA stacks, 0.94 $\pm$ 0.08 for LA 2CH, 0.94 $\pm$ 0.08 for LA 3CH, and 0.94 $\pm$ 0.07 for LA 4CH.}


\begin{table}[t]
\centering
\begin{tabular}{r c c c}
\toprule
\multicolumn{4}{c}{\textbf{Landmark Detection}}\\
\midrule
Localization Error & LA 2CH & LA 3CH & LA 4CH\\
\midrule
Apex 		        & 4.2 $\pm$ 2.5 & 4.5 $\pm$ 3.0 & 4.6 $\pm$ 3.1\\
Mitral Valve (Side I)  & 3.6 $\pm$ 2.6 & 3.5 $\pm$ 2.6 & 3.2 $\pm$ 2.3\\
Mitral Valve (Side II) & 3.9 $\pm$ 2.7 & 			  & 			  \\
\bottomrule
\end{tabular}
\caption{Landmark localization errors in mm (mean $\pm$ std).}
\label{tab:lms}
\end{table}

\begin{figure}[!t]
\centering
\includegraphics[width=0.85\columnwidth]{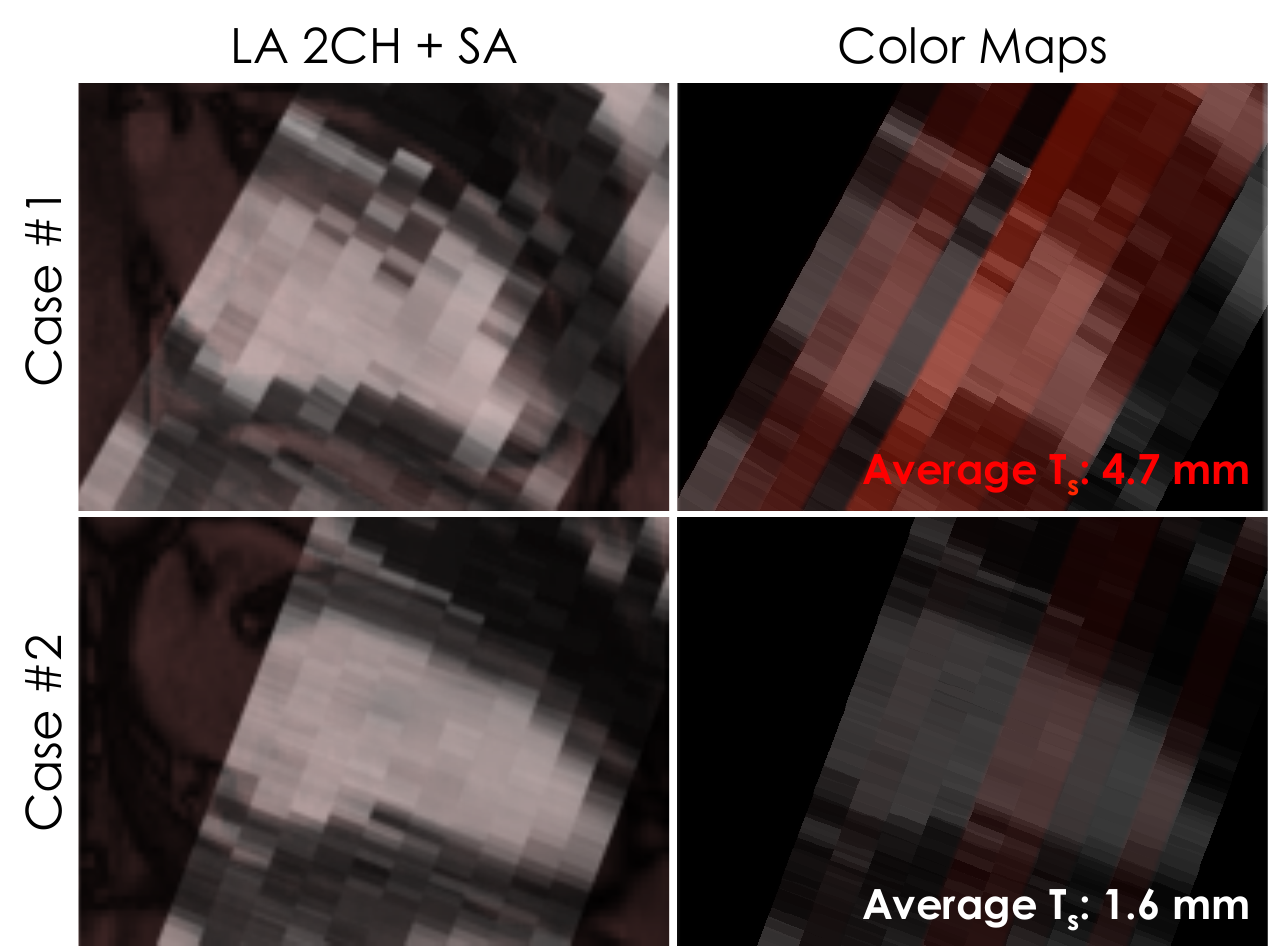}
\caption{Results for motion detection on UKBB in two cases, one with (case \#1) and one without motion corruption (case \#2). In the second column, the color maps of the translation magnitude for each slice are overlaid on top of the SA stacks.}
\label{fig:motion:results}
\end{figure}

\begin{figure}[!t]
\centering
\includegraphics[width=0.85\columnwidth]{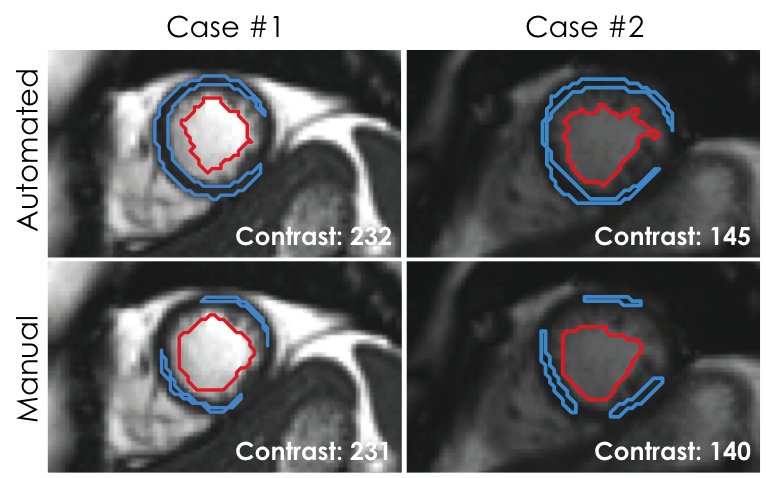}
\caption{Results for contrast estimation on UKBB in two cases, one with high (case \#1) and one with low contrast (case \#2). The ROIs from which the mean intensities are estimated are shown in red and cyan.}
\label{fig:contrast:results}
\end{figure}

First are reported the results on UKBB. For accuracy assessment of heart coverage estimation, 3 of the 3000 cases were excluded from the analysis: one due to the lack of LA images, and two for failing the sanity check on all the LA images. The ROC analysis performed on the remaining 2997 images returned an optimal threshold of 90\%. The results of the binary classification test are reported in Table \ref{tab:UKBB_coverage}. For accuracy assessment of motion detection, 3 of the 1500 cases were excluded from the analysis: one due to the lack of the SA stack and two for failing the sanity check. An ROC-like analysis was performed on the remaining 1497 images to select the thresholds $T_A$ and $T_B$. The results of the binary classification test, obtained for $T_A = 3.4$ mm and $T_B = 6$ mm, are reported in Table \ref{tab:UKBB_motion}. For accuracy assessment of contrast estimation, 3 of the 100 images were excluded from the analysis for failing the sanity check. Results for Pearson's correlation coefficient, linear regression and Bland-Altman analyses between automatically and manually estimated contrast values are reported in Table \ref{tab:UKBB_contrast} and in Fig. 11 (Supplementary Material). Examples of the results obtained for the three checks on UKBB are shown in Figs. \ref{fig:coverage:results}, \ref{fig:motion:results}, \ref{fig:contrast:results} and 10 (Supplementary Material).

\begin{my_mdframed}{Quality Control on UKBB}
\begin{table}[H]
\centering
\begin{tabular}{c c || c r | c c}
\toprule
\multicolumn{6}{c}{\textbf{Heart Coverage Estimation}}\\
\midrule
Sensitivity & Specificity &  & \multicolumn{1}{c}{} &  \multicolumn{2}{c}{Visual Assessment}\\
\midrule
\multirow{2}{*}{88\%} & \multirow{2}{*}{99\%}&  & Proposed & 49 (TP) & 15 (FP)\\
& &  & Technique & 7 (FN) & 2926 (TN)\\
\bottomrule
\end{tabular} 
\caption{Classification results for heart coverage estimation using a 90\% coverage threshold. Positive cases correspond to cases with insufficient coverage.}
\label{tab:UKBB_coverage}
\end{table}
\vspace{-5pt}

\begin{table}[H]
\centering
\begin{tabular}{c c || c r | c c}
\toprule
\multicolumn{6}{c}{\textbf{Inter-Slice Motion Detection}}\\
\midrule
Sensitivity & Specificity &  & \multicolumn{1}{c}{} &  \multicolumn{2}{c}{Visual Assessment}\\
\midrule
\multirow{2}{*}{85\%} & \multirow{2}{*}{95\%}&  & Proposed & 213 (TP) & 58 (FP)\\
& &  & Technique & 39 (FN) & 1187 (TN)\\
\bottomrule
\end{tabular} 
\caption{Classification results for motion detection using $T_A = 3.4$ mm and $T_B = 6$ mm. Positive cases correspond to motion-corrupted cases.}
\label{tab:UKBB_motion}
\end{table}
\vspace{-5pt}

\begin{table}[H]
\centering
\begin{tabular}{r c c c c c c}
\toprule
\multicolumn{7}{c}{\textbf{Cardiac Image Contrast Estimation}}\\
\midrule
& R & Bias & Std & a & b & Mean\\
\midrule
Original Images & 0.95 & -0.6 & 12.1 & 0.96 & 7.8 & 190\\
Normalized Images & 0.94 & -0.7 & 12.4 & 0.97 & 5.3 & 169\\
\bottomrule
\end{tabular}
\caption{Correlation coefficient (R), bias and std for Bland-Altman analysis, linear regression coefficients (a and b) and mean measured value between automatically and manually estimated cardiac image contrast, both on original images and after histogram normalization, in a.u..}
\label{tab:UKBB_contrast}
\end{table}
\end{my_mdframed}

\textclr{\blfootnote{AE.4,\\R2.8}Then are reported the results on UKDHP. For accuracy assessment of heart coverage estimation, all cases passed the sanity check. The ROC analysis returned an optimal threshold of 92\% coverage, and the results of the subsequent binary classification test are reported in Table \ref{tab:UKDHP_coverage}. For accuracy assessment of motion detection, 1 of the 100 cases was excluded from the analysis for failing the sanity check. The ROC-like analysis was performed on the remaining 99 images to select the thresholds $T_A$ and $T_B$. The results of the binary classification test, obtained for $T_A = 3$ mm and $T_B = 6$ mm, are reported in Table \ref{tab:UKDHP_motion}. For accuracy assessment of contrast estimation, 9 of the 100 images were excluded from the analysis for failing the sanity check. Results for Pearson's correlation coefficient, linear regression and Bland-Altman analyses between automatically and manually estimated contrast values are reported in Table \ref{tab:UKDHP_contrast} and in Fig. 11 (Supplementary Material).}

\begin{my_mdframed}{Quality Control on UKDHP}
\begin{table}[H]
\centering
\begin{tabular}{c c || c r | c c}
\toprule
\multicolumn{6}{c}{\textbf{Heart Coverage Estimation}}\\
\midrule
Sensitivity & Specificity &  & \multicolumn{1}{c}{} &  \multicolumn{2}{c}{Visual Assessment}\\
\midrule
\multirow{2}{*}{100\%} & \multirow{2}{*}{100\%}&  & Proposed & 5 (TP) & 0 (FP)\\
& &  & Technique & 0 (FN) & 95 (TN)\\
\bottomrule
\end{tabular} 
\caption{Classification results for heart coverage estimation using a 92\% coverage threshold. Positive cases correspond to cases with insufficient coverage.}
\label{tab:UKDHP_coverage}
\end{table}
\vspace{-5pt}

\begin{table}[H]
\centering
\begin{tabular}{c c || c r | c c}
\toprule
\multicolumn{6}{c}{\textbf{Inter-Slice Motion Detection}}\\
\midrule
Sensitivity & Specificity &  & \multicolumn{1}{c}{} &  \multicolumn{2}{c}{Visual Assessment}\\
\midrule
\multirow{2}{*}{78\%} & \multirow{2}{*}{90\%}&  & Proposed & 14 (TP) & 8 (FP)\\
& &  & Technique & 4 (FN) & 73 (TN)\\
\bottomrule
\end{tabular} 
\caption{Classification results for motion detection using $T_A = 3$ mm and $T_B = 6$ mm. Positive cases correspond to motion-corrupted cases.}
\label{tab:UKDHP_motion}
\end{table}
\vspace{-5pt}

\begin{table}[H]
\centering
\begin{tabular}{r c c c c c c}
\toprule
\multicolumn{7}{c}{\textbf{Cardiac Image Contrast Estimation}}\\
\midrule
& R & Bias & Std & a & b & Mean\\
\midrule
Original Images & 0.94 & -12.3 & 27.3 & 0.98 & -4.9 & 335\\
Normalized Images & 0.94 & -15.5 & 38.8 & 0.96 & 5.4 & 498\\
\bottomrule
\end{tabular}
\caption{Correlation coefficient (R), bias and std for Bland-Altman analysis, linear regression coefficients (a and b) and mean measured value between automatically and manually estimated cardiac image contrast, both on original images and after histogram normalization, in a.u..}
\label{tab:UKDHP_contrast}
\end{table}
\end{my_mdframed}

\section{Discussion}

The results obtained for the landmark localization experiment show that the average localization error is around 3.9 mm (roughly two pixels) and is thus small compared to the reconstructed slice thickness in both datasets (10 mm), suggesting the reliability of the landmark detection technique for the sake of heart coverage estimation. \textclr{\blfootnote{R1.2}The proposed hybrid decision forest method is based upon a previous implementation for landmark detection \cite{Oktay2017} which consisted of a multi-stage approach devised to increase the robustness to large variations in distances and orientation of the landmarks. It is worth mentioning that initial experiments performed using this approach showed no measurable improvement with respect to the single-stage one (perhaps due to the size of the training set and to the consistency of the orientation of the images), which was thus preferred (Fig. 12, Supplementary Material).} \textclr{\blfootnote{R4.7,\\R2.8}The high DSC values obtained for the PSMs suggest their reliability for both motion detection and contrast estimation. The fact that the PSMs of the SA stacks are slightly worse than those of the LA images (0.90 vs 0.94) is mainly due to a lower response of the model in the apical slices, where a different thresholding value would have been beneficial. However, this does not cause a direct problem on the proposed pipeline, which never thresholds PSMs for segmentation purposes and instead exploits their probabilistic nature.}

The first set of experiments involving the whole pipeline was aimed at assessing its accuracy on UKBB. The binary classification test on coverage estimation performed on 2997 cases from UKBB indicates the high accuracy of the proposed technique, with sensitivity = 88\% and specificity = 99\%. The interpretation of these results is hindered by the strong class imbalance between cases with sufficient and insufficient coverage, and thus a more detailed analysis of the reported confusion matrix is required. By applying the proposed automated technique, it is possible to correctly detect 88\% of the cases with insufficient coverage, and thus to lower the percentage of undetected wrongly imaged cases from 1.9\% to 0.2\%. This comes at the price of having to visually check an additional 0.5\% of cases that actually featured a sufficient coverage. Notably, several of the 15 FP cases actually had a sub-optimal coverage, but not of the amount required to be considered as wrongly imaged following the criterion adopted during visual inspection. Compared to our previous work \cite{Tarroni2017a}, the present approach makes use of three LA images instead of just one. The redundancy offered by exploiting all the available LA views allows a more robust and reliable estimation: this is suggested by the higher sensitivity and specificity achieved (88\% vs 73\% and 99\% vs 98\%, respectively, although a direct comparison is not completely fair since the UKBB subset used in \cite{Tarroni2017a} was different from the present one) and by the lower number of cases excluded due to failing the sanity check (down from 89 to 3). Of note, this check is able to indirectly detect and exclude LA images with high noise levels, wrong acquisition planning or wrong file naming that make the landmark localization unreliable, and in the present implementation only cases in which all the three LA views yielded bad landmark detection had to be excluded from the coverage assessment. \textclr{\blfootnote{R1.10,\\R4.8,\\R4.3}Zhang et al. \cite{Zhang2016} addressed coverage estimation by performing fully-supervised CNN-based slice classification to detect stacks with missing basal (MBS) or apical slices (MAS). In their later work \cite{Zhang2017}, they acknowledged the need for a large amount of labelled data during training to achieve good generalization: to mitigate this issue, the authors have increased the size of the training set using generative networks (reaching average accuracies of 93\% for MAS and 89\% for MBS on a dataset of 3400 cases from UKBB). The use of different subsets of data from the UKBB and the different validation strategies (detection of missing slices separately in the apical and in the basal region vs detection of overall non-optimal cases) make the comparison between the two approaches not straightforward. The main advantage of the approach of Zhang et al. is that it can detect problematic scans using only the SA stack, while our pipeline relies on the presence of at least one of the LA views (which are, however, routinely acquired in most CMR protocols). On the other hand, we believe there is a clinical and practical advantage in measuring the relative coverage instead of performing binary classification: cases with only slightly sub-optimal coverage could still be included in the following analyses, especially when the lack of coverage is in the apical area. Moreover, while their approach completely relies on feature extraction from single slices and thus small image perturbations can potentially lead to misclassification, our approach is designed to exploit the redundancy offered by the multiple LA views for greater robustness.}

The reported results on UKBB for motion detection indicate that the proposed approach achieves sensitivity = 85\% and specificity = 95\% over 1497 cases. By applying the proposed automated technique, it is possible to lower the percentage of undetected motion-corrupted cases from 16.8\% to 2.6\%. This comes at the price of having to visually check 3.9\% cases that were visually deemed motion-free. It is worth to note that the binary classification of stacks based on the visual assessment of motion is a difficult task in itself, limiting the measurable accuracy of any technique. A more thorough examination would require a slice-by-slice visual classification, which is however impractical for datasets of this size.

The accuracy of the contrast estimation technique on UKBB is indicated by very high correlation coefficients and regression lines near unity both for images before and after contrast normalization. Bland-Altman analyses show negligible biases and narrow limits of agreement with respect to the mean measured values, suggesting the high accuracy of the technique.

\textclr{\blfootnote{AE.4,\\R2.8}The second set of experiments was aimed at assessing the accuracy of the pipeline (trained on UKBB) on the UKDHP dataset, thus testing its generalization properties, and yielded encouraging results. Regarding heart coverage estimation, our technique was able to correctly identify all sub-optimal cases. Regarding motion detection, it returned slightly lower values for sensitivity and specificity than those obtained on UKBB: while this might be due to a lower accuracy of the extracted PSMs, we noted that motion in the UKDHP dataset is considerably less pronounced than on UKBB, so it is easier to misclassify borderline cases. Regarding contrast estimation, the technique showed again very high correlation coefficients and regression lines near unity. The increased difficulty in dealing with a testing dataset different from the training one can be seen in the slightly higher number of cases failing the sanity check (up from 3 to 9) and in bigger biases, still however negligible when compared to the mean measured values. In general, the small size of the UKDHP dataset should be taken in consideration when evaluating these results, especially for binary classification tests where the misclassification of a single case can have a very large influence on the accuracy figures. However, we believe the reported results show that the proposed approach generalizes well to previously unseen datasets, coping with differences in the acquisition protocols.}

Our approach to quality control does not attempt to directly classify sub-optimal cases for two reasons. First, this allows the complete circumvention of any class-imbalance issues, since the only learning-based portions of our pipeline aim at the identification of structures that are present in every image. Second, the implemented pipeline does not constitute a ``black-box" approach: each quality check produces quantitative metrics with a clear meaning, which can be of great value in informing the MR operators on the type and the entity of the identified issues. \textclr{\blfootnote{AE.1,\\R1.2}Importantly, the proposed pipeline could be adopted also using different techniques for landmark detection and probabilistic segmentation. One major requirement for these alternative methods would be the generation of fuzzy segmentations maps providing a probabilistic representation of the target structures: this allows the assessment of their reliability for both motion detection and contrast estimation, otherwise  unfeasible with standard, hard segmentations.}

The main limitation affecting our approach is that \textclr{\blfootnote{R2.2}no quality check is performed on the manual selection of the imaging planes for LA and SA images}, which can be subject to error. However, countermeasures have been implemented to deal with this issue. Regarding coverage estimation, the redundancy offered by exploiting all the three LA views and the adoption of a sanity check helps to minimize the issue. Regarding motion detection, a slighty off-axis LA image still correctly represents the cardiac anatomy, and the initial 3D registration step will position it correctly with respect to the SA stack.

\section{Conclusion}
In this paper, a fully-automated, learning-based pipeline for quality control of CMR images has been presented. The implemented quality checks are heart coverage estimation, inter-slice motion detection and cardiac image contrast estimation for short-axis image stacks. The pipeline uses hybrid random forests to extract probabilistic segmentation maps and identify landmarks on long- and short-axis images, and then leverages these information to perform the quality checks. It was tested on up to 3000 cases from the UKBB \textclr{\blfootnote{AE.4,\\R2.8}as well as on 100 cases from the UKDHP}, and compared to the results of visual or manual analyses to evaluate its accuracy. The results suggest that the proposed approach is able to perform the quality checks with a high accuracy across different datasets. With the recent launch of several initiatives for the acquisition of large-scale CMR datasets, there is a strong need for robust quality control tools in order to facilitate and ensure the reliability of the analyses performed as part of clinical studies. In addition, the low computational time required by the proposed pipeline makes it potentially deployable at the acquisition site, allowing the almost real-time assessment of the scan and the potential triggering of a new acquisition.

\section*{Acknowledgments}

This research has been conducted using the UK Biobank Resource \cite{Petersen2016} under Application Number 18545. The first author benefited from a Marie Skodowska-Curie Fellowship. \textclr{The authors acknowledge funding by EPSRC programme (EP/P001009/1), British Heart Foundation (NH/17/1/32725) and support by NIHR Imperial Biomedical Research Centre.}

\bibliographystyle{IEEEtran}

\bibliography{qc_journal_arxiv}

\begin{thebibliography}{10}
\providecommand{\url}[1]{#1}
\csname url@samestyle\endcsname
\providecommand{\newblock}{\relax}
\providecommand{\bibinfo}[2]{#2}
\providecommand{\BIBentrySTDinterwordspacing}{\spaceskip=0pt\relax}
\providecommand{\BIBentryALTinterwordstretchfactor}{4}
\providecommand{\BIBentryALTinterwordspacing}{\spaceskip=\fontdimen2\font plus
\BIBentryALTinterwordstretchfactor\fontdimen3\font minus
  \fontdimen4\font\relax}
\providecommand{\BIBforeignlanguage}[2]{{%
\expandafter\ifx\csname l@#1\endcsname\relax
\typeout{** WARNING: IEEEtran.bst: No hyphenation pattern has been}%
\typeout{** loaded for the language `#1'. Using the pattern for}%
\typeout{** the default language instead.}%
\else
\language=\csname l@#1\endcsname
\fi
#2}}
\providecommand{\BIBdecl}{\relax}
\BIBdecl

\bibitem{Zhuo2006}
J.~Zhuo and R.~P. Gullapalli, ``{MR Artifacts, Safety, and Quality Control},''
  \emph{RadioGraphics}, vol.~26, no.~1, pp. 275--297, jan 2006.

\bibitem{Ferreira2013}
P.~F. Ferreira, P.~D. Gatehouse, R.~H. Mohiaddin, and D.~N. Firmin,
  ``{Cardiovascular magnetic resonance artefacts},'' \emph{Journal of
  Cardiovascular Magnetic Resonance}, vol.~15, no.~1, p.~41, 2013.

\bibitem{Petersen2016}
S.~E. Petersen, P.~M. Matthews, J.~M. Francis, M.~D. Robson, F.~Zemrak,
  R.~Boubertakh, A.~A. Young, S.~Hudson, P.~Weale, S.~Garratt, R.~Collins,
  S.~Piechnik, and S.~Neubauer, ``\BIBforeignlanguage{En}{{UK Biobank's
  cardiovascular magnetic resonance protocol.}}''
  \emph{\BIBforeignlanguage{En}{Journal of Cardiovascular Magnetic Resonance}},
  vol.~18, no.~1, p.~8, jan 2016.

\bibitem{Klinke2013}
V.~Klinke, S.~Muzzarelli, N.~Lauriers, D.~Locca, G.~Vincenti, P.~Monney, C.~Lu,
  D.~Nothnagel, G.~Pilz, M.~Lombardi, A.~C. van Rossum, A.~Wagner, O.~Bruder,
  H.~Mahrholdt, and J.~Schwitter, ``{Quality assessment of cardiovascular
  magnetic resonance in the setting of the European CMR registry: description
  and validation of standardized criteria},'' \emph{Journal of Cardiovascular
  Magnetic Resonance}, vol.~15, no.~1, p.~55, jun 2013.

\bibitem{Coupe2010}
P.~Coup{\'{e}}, J.~V. Manj{\'{o}}n, E.~Gedamu, D.~Arnold, M.~Robles, and D.~L.
  Collins, ``{Robust Rician noise estimation for MR images},'' \emph{Medical
  Image Analysis}, vol.~14, no.~4, pp. 483--493, 2010.

\bibitem{Maximov2012}
I.~I. Maximov, E.~Farrher, F.~Grinberg, and N.~{Jon Shah}, ``{Spatially
  variable Rician noise in magnetic resonance imaging},'' \emph{Medical Image
  Analysis}, vol.~16, no.~2, pp. 536--548, 2012.

\bibitem{Gedamu2008}
E.~L. Gedamu, D.~L. Collins, and D.~L. Arnold, ``{Automated quality control of
  brain MR images},'' \emph{Journal of Magnetic Resonance Imaging}, vol.~28,
  no.~2, pp. 308--319, aug 2008.

\bibitem{UKDHP}
``{Digital Heart Project},'' in \emph{https://digital-heart.org}.

\bibitem{Alba2018}
X.~Alb{\`{a}}, K.~Lekadir, M.~Perea{\~{n}}ez, P.~Medrano-Gracia, A.~A. Young,
  and A.~F. Frangi, ``{Automatic initialization and quality control of
  large-scale cardiac MRI segmentations.}'' \emph{Medical Image Analysis},
  vol.~43, pp. 129--141, jan 2018.

\bibitem{Zhang2016}
L.~Zhang, A.~Gooya, B.~Dong, R.~Hua, S.~E. Petersen, P.~Medrano-Gracia, and
  A.~F. Frangi, ``{Automated Quality Assessment of Cardiac MR Images Using
  Convolutional Neural Networks},'' in \emph{SASHIMI}.\hskip 1em plus 0.5em
  minus 0.4em\relax Springer, Cham, oct 2016, vol. LNCS, no. 9968, pp.
  138--145.

\bibitem{Zhang2017}
L.~Zhang, A.~Gooya, and A.~F. Frangi, ``{Semi-supervised Assessment of
  Incomplete LV Coverage in Cardiac MRI Using Generative Adversarial Nets
  Chapter},'' \emph{SASHIMI}, vol. LNCS, pp. 138--145, 2017.

\bibitem{Tarroni2017a}
G.~Tarroni, O.~Oktay, W.~Bai, A.~Schuh, H.~Suzuki, J.~Passerat-Palmbach,
  B.~Glocker, A.~de~Marvao, D.~O'Regan, S.~Cook, and D.~Rueckert,
  ``{Learning-Based Heart Coverage Estimation for Short-Axis Cine Cardiac MR
  Images},'' \emph{FIMH}, vol. LNCS, no. 10263, pp. 73--82, jun 2017.

\bibitem{McClelland2013}
J.~R. McClelland, D.~J. Hawkes, T.~Schaeffter, and A.~P. King, ``{Respiratory
  motion models: a review.}'' \emph{Medical Image Analysis}, vol.~17, no.~1,
  pp. 19--42, jan 2013.

\bibitem{Lotjonen2004a}
J.~L{\"{o}}tj{\"{o}}nen, M.~Pollari, S.~Kivist{\"{o}}, and K.~Lauerma,
  ``{Correction of Movement Artifacts from 4-D Cardiac Short- and Long-Axis MR
  Data},'' \emph{MICCAI}, vol. LNCS, no. 3217, pp. 405--412, sep 2004.

\bibitem{Oktay2016}
O.~Oktay, G.~Tarroni, W.~Bai, A.~de~Marvao, D.~P. O'Regan, S.~A. Cook, and
  D.~Rueckert, ``{Respiratory motion correction for 2D cine cardiac MR images
  using probabilistic edge maps},'' in \emph{Computing in Cardiology Conference
  (CinC)}.\hskip 1em plus 0.5em minus 0.4em\relax IEEE, 2016, pp. 129--132.

\bibitem{Sinclair2017}
M.~Sinclair, W.~Bai, E.~Puyol-Ant{\'{o}}n, O.~Oktay, D.~Rueckert, and A.~P.
  King, ``{Fully Automated Segmentation-Based Respiratory Motion Correction of
  Multiplanar Cardiac Magnetic Resonance Images for Large-Scale Datasets},''
  \emph{MICCAI}, vol. LNCS, Part, no. 10434, pp. 332--340, sep 2017.

\bibitem{Yang2017}
D.~Yang, P.~Wu, C.~Tan, K.~M. Pohl, L.~Axel, and D.~Metaxas, ``{3D Motion
  Modeling and Reconstruction of Left Ventricle Wall in Cardiac MRI},'' in
  \emph{FIMH}, vol. LNCS 10263.\hskip 1em plus 0.5em minus 0.4em\relax
  Springer, Cham, jun 2017, pp. 481--492.

\bibitem{McLeish2002}
K.~McLeish, D.~L.~G. Hill, D.~Atkinson, J.~M. Blackall, and R.~Razavi, ``{A
  study of the motion and deformation of the heart due to respiration.}''
  \emph{IEEE Transactions on Medical Imaging}, vol.~21, no.~9, pp. 1142--50,
  sep 2002.

\bibitem{Dietrich2007}
O.~Dietrich, J.~G. Raya, S.~B. Reeder, M.~F. Reiser, and S.~O. Schoenberg,
  ``{Measurement of signal-to-noise ratios in MR images: Influence of
  multichannel coils, parallel imaging, and reconstruction filters},''
  \emph{Journal of Magnetic Resonance Imaging}, vol.~26, no.~2, pp. 375--385,
  aug 2007.

\bibitem{Aja-Fernandez2014}
S.~Aja-Fern{\'{a}}ndez, G.~Vegas-S{\'{a}}nchez-Ferrero, and
  A.~Trist{\'{a}}n-Vega, ``{Noise estimation in parallel MRI: GRAPPA and
  SENSE},'' \emph{Magnetic Resonance Imaging}, vol.~32, no.~3, pp. 281--290,
  apr 2014.

\bibitem{Wolff1997}
S.~D. Wolff and R.~S. Balaban, ``{Assessing contrast on MR images.}''
  \emph{Radiology}, vol. 202, no.~1, pp. 25--29, jan 1997.

\bibitem{Oktay2017}
O.~Oktay, W.~Bai, R.~Guerrero, M.~Rajchl, A.~de~Marvao, D.~P. O'Regan, S.~A.
  Cook, M.~P. Heinrich, B.~Glocker, and D.~Rueckert, ``{Stratified Decision
  Forests for Accurate Anatomical Landmark Localization in Cardiac Images},''
  \emph{IEEE Transactions on Medical Imaging}, vol.~36, no.~1, pp. 332--342,
  jan 2017.

\bibitem{Criminisi2011}
A.~Criminisi, J.~Shotton, and E.~Konukoglu, ``{Decision Forests: A Unified
  Framework for Classification, Regression, Density Estimation, Manifold
  Learning and Semi-Supervised Learning},'' \emph{Foundations and
  Trends{\textregistered} in Computer Graphics and Vision}, vol.~7, no. 2-3,
  pp. 81--227, 2011.

\bibitem{Gall2011}
J.~Gall, A.~Yao, N.~Razavi, L.~{Van Gool}, and V.~Lempitsky, ``{Hough Forests
  for Object Detection, Tracking, and Action Recognition},'' \emph{IEEE
  Transactions on Pattern Analysis and Machine Intelligence}, vol.~33, no.~11,
  pp. 2188--2202, nov 2011.

\bibitem{Dollar2015}
P.~Dollar and C.~L. Zitnick, ``{Fast Edge Detection Using Structured
  Forests},'' \emph{IEEE Transactions on Pattern Analysis and Machine
  Intelligence}, vol.~37, no.~8, pp. 1558--1570, 2015.

\bibitem{Criminisi2013}
A.~Criminisi, D.~Robertson, E.~Konukoglu, J.~Shotton, S.~Pathak, S.~White, and
  K.~Siddiqui, ``{Regression forests for efficient anatomy detection and
  localization in computed tomography scans},'' \emph{Medical Image Analysis},
  vol.~17, no.~8, pp. 1293--1303, 2013.

\bibitem{Celeux1996}
G.~Celeux and G.~Soromenho, ``{An entropy criterion for assessing the number of
  clusters in a mixture model},'' \emph{Journal of Classification}, vol.~13,
  no.~2, pp. 195--212, sep 1996.

\bibitem{Bai2017a}
W.~Bai, M.~Sinclair, G.~Tarroni, O.~Oktay, M.~Rajchl, G.~Vaillant, A.~M. Lee,
  N.~Aung, E.~Lukaschuk, M.~M. Sanghvi, F.~Zemrak, K.~Fung, J.~M. Paiva,
  V.~Carapella, Y.~J. Kim, H.~Suzuki, B.~Kainz, P.~M. Matthews, S.~E. Petersen,
  S.~K. Piechnik, S.~Neubauer, B.~Glocker, and D.~Rueckert, ``{Automated
  cardiovascular magnetic resonance image analysis with fully convolutional
  networks},'' \emph{Journal of Cardiovascular Magnetic Resonance}, oct 2017.

\bibitem{Nyul2000}
L.~G. Ny{\'{u}}l, J.~K. Udupa, and X.~Zhang, ``{New variants of a method of MRI
  scale standardization},'' \emph{IEEE Transactions on Medical Imaging},
  vol.~19, no.~2, pp. 143--150, 2000.

\end{thebibliography}

\end{document}